# Adversarial Attack for Uncertainty Estimation: Identifying Critical Regions in Neural Networks

Ismail Alarab[1] · Simant Prakoonwit[1]

## Abstract
We propose a novel method to capture data points near decision boundary in neural network that are often referred to a specific type of uncertainty. In our approach, we sought to perform uncertainty estimation based on the idea of adversarial attack method. In this paper, uncertainty estimates are derived from the input perturbations, unlike previous studies that provide perturbations on the model's parameters as in Bayesian approach. We are able to produce uncertainty with couple of perturbations on the inputs. Interestingly, we apply the proposed method to datasets derived from blockchain. We compare the performance of model uncertainty with the most recent uncertainty methods. We show that the proposed method has revealed a significant outperformance over other methods and provided less risk to capture model uncertainty in machine learning.

## 1 Introduction

Standard Neural network models have admitted great success in approximating functions that map the inputs to desirable outputs in several domains [1–3]. Nevertheless, a single function approximation often leads to overconfident and erroneous predictions. To reliably perform predictions, uncertainty estimation is a desirable approach to provide on which inputs the model is not certain about its predictions. Uncertainty estimation has increasingly received considerable attention in the recent years. Several studies are conducted to express uncertainty such as Monte Carlo dropout (MC-dropout) [4], Deterministic Uncertainty Quantification (DUQ) [5] and Deterministic Uncertainty Estimation (DUE) [6]. However, these methods are unable to capture noisy data points between the overlapping class distributions (critical region) due to their inherent approximations. For instance, the perturbations of the decision

---

✉ Ismail Alarab
  ialarab@bournemouth.ac.uk

1  Bournemouth University, Poole, UK

   

boundary given by MC-dropout can only cause fluctuations in the model output by the inputs lying in this critical region (decision boundary zone). However, a noisy data point lying completely in a different region hinders the performance of MC-dropout wherein datapoints falling in wrong classes cannot be triggered by the variability of the decision boundary. This issue is further illustrated in the upcoming sections. Generally, existing methods relies on weight variability or function approximation variability which in effect cause perturbations in the decision boundary.

This behaviour hinders the performance of these approaches since they cannot capture noisy data points in overlapping regions leading to high number of misclassified and certain predictions. In this paper, we propose a new method based on adversarial attack to estimate uncertainties in neural network models based on binary classification task. We refer to our proposed method as Monte Carlo based Adversarial Attack abbreviated by "MC-AA". In this paper, "Critical region" is inspired from the known hypothesis test in statistics wherein the null hypothesis is accepted or rejected. By analogy, classification in machine learning resembles as hypothesis test to perform decision-making, wherein critical region means the region between the opposite classes or the decision boundary zone.

Initially, adversarial attacks/examples are defined as the inputs to machine learning that an attacker has introduced to fool the outputs of the model [7–9]. These attacks have a great impact on the security and integrity of machine learning model resulting in poor decision-making. Motivated by previous work, we estimate uncertainty of the predictions using adversarial attack. Firstly, we train a standard neural network then we apply MC-AA method during the testing phase so that multiple outputs on each test point are introduced. Hence, we compute mutual information on these outputs to estimate the uncertainty. Unlike previous studies that apply perturbations to the model parameters (e.g. in Bayesian approach), our method sought to estimate uncertainty by perturbing the relevant inputs in a guided way. The uncertainty estimates are highly tied with the location of the test point with respect to the decision boundary. As a result, a perturbed input lying near decision boundary produces fluctuated outputs between different classes. Furthermore, an input lying in overlapping regions is enforced to move back and forth between the different classes leading to uncertain predictions using MC-AA. Consequently, the proposed method is likely to consider any point lying on the border of the class distributions as uncertain. Accordingly, this reduces the risk to produce wrong and certain predictions. The proposed method is backed up by the results to obtain uncertainties and support our main idea.

The contributions of this paper are as follows:

- We show that adversarial attack is able to capture model uncertainty in binary classification tasks.
- We demonstrate a well-defined way to estimate and evaluate the uncertainty of neural network model using MC-AA. In addition, we illustrate the behaviour of the proposed method in the feature space.
- We introduce the relevance of our proposed method to Bayesian approximations under given conditions.
- We perform a comparative analysis between the most recent uncertainty methods using datasets derived from blockchain. Our method shows superior success over the previous work to capture uncertainty.

The following sections are organised as follows: Sect. 2 provides the overview of the related work. Section 3 demonstrates the background of the proposed method that is detailed in Sect. 4. Section 5 provides the model uncertainty measurements using the proposed method. The experiments are provided in Sect. 6 and discussed in Sect. 7. A conclusion is stated in Sect. 8.

## 2 Overview of Related Work

Knowing what the model is uncertain about is a desirable interest. Bayesian neural network (BNN) has admitted great success in model uncertainty [10]. This approach introduces priors over the weights of the neural network in order to find their posterior distributions; however, this is computationally expensive. Several studies have been conducted to tackle the time complexity problem of BNN by approximating the posterior distributions of the model, such as the case in variational inference [11] and Markov Chain Monte Carlo with Hamiltonian Dynamics [12]. However, they are hard to scale to large datasets [13]. Recent studies have proposed more efficient methods to estimate uncertainties in neural networks known as Monte-Carlo dropout (MC-dropout) [4], Deterministic Uncertainty Quantification (DUQ) [5], and Deterministic Uncertainty Estimation (DUE) [6]. MC-dropout is equivalent to probabilistic Bayesian approximations to capture model uncertainty [4]. This method is derived from multiple stochastic forward passes on a trained neural network with activated dropout during the testing phase to estimate uncertainties. In MC-dropout, the points falling near the decision boundary are more likely to be uncertain about [14]. Nevertheless, this method reveals two drawbacks. It requires many stochastic forward passes to output stable estimates, and its uncertainty estimates are not reliable due to the high number of erroneous and certain data points. DUQ is based on the idea of Radial Basis Function (RBF) network incorporated with gradient penalty, wherein its output is squashed with Gaussian function to quantify the distance from the tested point to the centroids of the given class distributions [5]. This method is good at finding Out-of-Distribution (OoD) data. However, it is unreliable to capture noisy data points where the distance is not a sufficient metric to convey uncertainty. DUE is based on the approximation of Gaussian process that scales to high dimensional data using variational inducing points with feature extractor of Deep Kernel Learning (DKL) [6]. Although its efficiency in capturing uncertainty, this method is not able to capture data points lying in between the overlapping class distributions. As DUE method is based on Gaussian process, this approach has never scaled to high dimensional datasets because of a lack of well performing kernel function [5]. In addition, a comprehensive review of most uncertainty methods in [15] as revealed the vigilance of previous studies on failures of model uncertainty caused by adversarial attacks. Also, the methods in [16, 17] has sought to improve the robustness of adversarial examples in machine learning. However, none of them has straightforwardly used the idea of adversarial attack to estimate uncertainty to the best of our knowledge.

## 3 Background

Adversaries are crafted perturbations of the legitimate inputs. The perturbed inputs influence the predictions of a trained model to output incorrect predictions [18]. Such inputs are designed by attackers to affect the security and integrity of the model. Some of these attacks incorporate white-box models meaning that the attacker acquires a detailed knowledge of the model's parameters and architecture and tries to design a perturbed input using the well-known Fast Gradient Sign Method (FGSM).

### 3.1 FGSM

FGSM is based on the gradient of the loss function with respect to the initial inputs. Consider a neural network incorporating a set of parameters $W = \{W_1, \ldots, W_L\}$ where $L$ is the number of layers. Consider a dataset with size $N$ as $X = \{x_1, \ldots, x_N\}$ and $Y = \{y_1, \ldots, y_N\}$, then the loss function of this model can be written as: $J(x, y)$. Hence, FGSM can be reformulated as follows:

$$x_{Adv} = x + \varepsilon.sign(\nabla_x J(x, y)), \quad (1)$$

where $x_{Adv}$ is the adversarial example, $\epsilon$ is a small number and $\nabla_x$ is the gradient with respect to the input, $x \in X$ and $y \in Y$. Basically, FGSM adds noise to the input data in the direction to the nearest decision boundary and scaled with $\epsilon$ as shown in Fig. 1. By providing the incorrect class to the loss function, FGSM enforces the data points to walk towards the decision boundary in the direction of the incorrect class. This type of attacks is rather tackled by adversarial training.

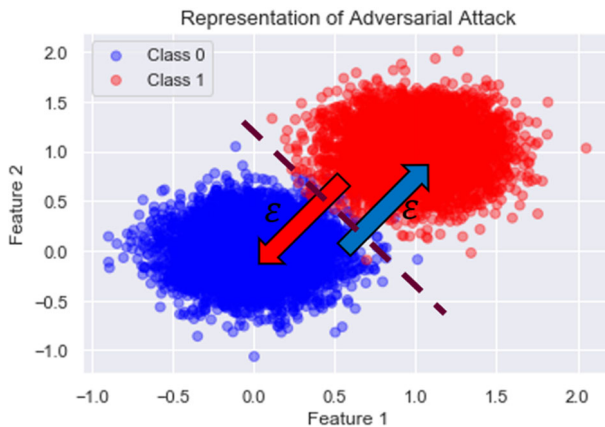

**Fig. 1** Toy example of adversarial attack with a linear classifier

### 3.2 Adversarial Training

Adversarial training increases the model robustness by learning adversarial examples over the train set [20]. This is also referred to data augmentation method so that the decision boundary is further extended to these sensitive perturbations. The robust model is obtained either by feeding the augmented data [19] or by minimisation of the modified objective loss function [9] that can be expressed as:

$$\tilde{J}(x, y) = \alpha J(x, y) + (1 - \alpha) J(x + \varepsilon.sign(\nabla_x J(x, y)), y), \quad (2)$$

where $\alpha$ is a regularisation parameter between 0 and 1, and other notations similar to that in Eq. 1. The modified objective function allows to minimax the loss with the consideration of the adversaries.

### 3.3 Adversarial Attacks and Uncertainty

The emergent uncertainty studies have discussed the impact of adversarial attacks on model uncertainty. For instance, Smith et al. [20] has demonstrated the mode failures of MC-dropout that is caused by adversaries. These failures are derived from the overconfident and certain predictions that cannot be captured by dropout. The review in [15] has revealed the awareness of previous studies regarding adversarial attacks in uncertainty. For instance, Bradshaw et al. [21] has proposed a hybrid model of GP and DNN which is robust to adversarial attacks to produce uncertainties. Pawlowski et al. [22] has introduced a new technique of variational approximation that produced competitive accuracies and robustness against adversarial attacks. Unlike previous studies, we perform uncertainty estimation directly by using adversarial attack method with random class label assumption on the whole data points. This is demonstrated in the upcoming section.

## 4 Method

Mainly, the uncertainty of the model falls into two major categories either epistemic or aleatoric uncertainty. Epistemic uncertainty underlies the new instances that are not learned yet by a model. Aleatoric is the noisy observations such as data lying between the overlapping classes. This type of noise is irreducible by acquiring more data in contrast to epistemic uncertainty. In this paper, we intend to capture uncertainty using adversarial attacks on the test set. For every data point, this is performed by feeding the neural network with multiple perturbed versions of the input linearly back and forth in the direction of the decision boundary as illustrated in Fig. 2. Hence, multiple outputs are produced which inform the tendency of a given data point, after perturbations, to fall in the opposite class. The variants of the produced output introduce the uncertainty on the given point using mutual information. The multiple perturbations are the multiple scaled gradients added to the input data as a function of epsilon. However, in order to apply FGSM, we need the target class in order to maximise or minimise the loss function. Thus, we need to arbitrarily assign a class to compute the gradients.

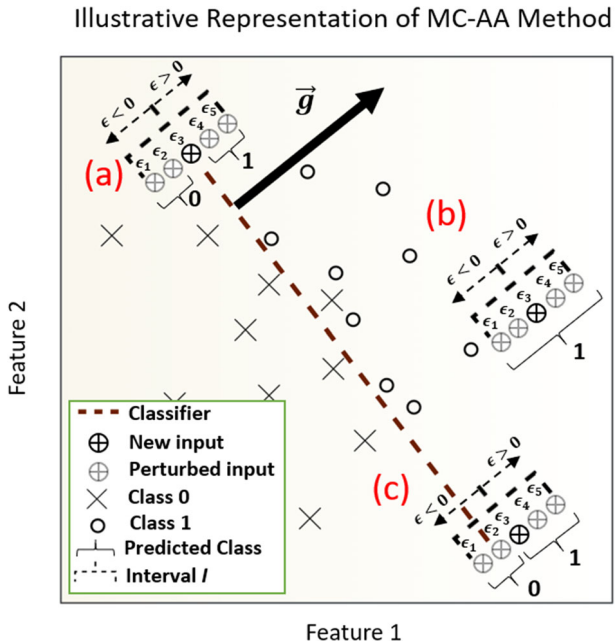

**Fig. 2** Illustrative representation of MC-AA method. Case (a) describes an input lying on the decision boundary. Case (b) describes an input lying completely in its corresponding class 1. Case (c) describes an input with its actual label 0 that lies in the region of class 1 (overlapping region). $\vec{g}$ represents the gradient of the loss function with respect to the input

### 4.1 Procedure of MC-AA

Adversarial examples are computed by feeding input data to the neural network and assigning a target class in the loss function. Then, the gradient of the loss function with respect to input is scaled and added as mentioned in Eq. 1. To perform MC-AA, we apply the following procedure:

1. Assume that all the test set belongs to the same class, e.g. class 0.
2. Perform multiple FGSM on each input for all values of $\varepsilon \in I = \{\varepsilon_{min}, \varepsilon_{min} + \beta, \ldots, 0, 0 + \beta, \ldots, \varepsilon_{max}\}$ with $\varepsilon_{max} = -\varepsilon_{min}$ as provided in Algorithm 1, where the given interval is evenly spaced by $\beta$ and symmetric around zero. One should note that standardisation of the dataset is a necessary step.
3. Compute mutual information metric to introduce uncertainty on each data point.

---

**Algorithm 1** MC-AA

**Require:**
- $\mathcal{M}$: is a trained neural network that maps the feature vectors to binary predictions (0 or 1).
- $J(.,.)$: is a loss function that takes a data point and its labels as inputs.
- **Input:**
  - **I**: is a discrete interval comprising different values of $\epsilon$. $\epsilon_{max}$ is a tuned hyper-parameter.
  - $x$: is a feature vector to be tested.
  - $\tilde{y}$: is an arbitrary output assumed on the input. Here, we assume that all test points are labelled $\tilde{y} = 0$.
- **Output:**
  - $\hat{y}$: is a set of outputs for a single data point. $\hat{y} = \{\hat{y}_{\epsilon_1}, \ldots, \hat{y}_{\epsilon_n}\}$, where $\epsilon_i \in I$.

**Function**

Compute the gradients of the loss function with respect to x as:
grad = $\nabla_x J(x, \tilde{y})$
**for all** $\epsilon_i \in I$ **do**

$\quad x_{\epsilon_i} = x + \epsilon_i . sign(grad);$
$\quad \hat{y}_{\epsilon_i} = \mathcal{M}(x_{\epsilon_i});$
$\quad$ return $\hat{y}_{\epsilon_i}$

**end for**

$\hat{y} = \{\hat{y}_{\epsilon_1}, \ldots, \hat{y}_{\epsilon_n}\};$
return $\hat{y}$

---

If the assumed class is equal to the actual one and $\epsilon > 0$, the output will maximise the loss function so that the perturbed instance will shift towards the incorrect class. In contrast, if $\varepsilon < 0$, the output will minimise the loss function and the perturbed instance will shift towards the correct class. Otherwise, the opposite of the preceding statements is true. As a result, the inputs falling near the decision boundaries are firstly targeted to provide fluctuating outputs with various values of $\varepsilon$, and the model classifies them as uncertain referring to Fig. 2. We further note the data point falling in the overlapping region of different classes as the case (c) in Fig. 2; MC-AA is able to capture such points, unlike MC-dropout method. MC-dropout performs slight variations on the decision boundary. As this method is based on stochastic forward passes, the variations on the decision boundary may not be able to move around misclassified instances near the decision boundary with highly trained instances of a similar class. In other words, the decision boundary is only able to perform variations with dropout in a critical region. instances with misclassified predictions are more likely to occur nearby decision boundary. While, MC-AA enforces datapoints to move back and forth in the direction of the decision boundary regardless of region. This leads any instance near the decision boundary to move between different classes, hence reflecting good uncertainty estimates. This is done at the cost of capturing some correct predictions with uncertainty, but they do not affect the performance of model uncertainty which is favoured. Also, the interval of epsilon is chosen symmetric and evenly spaced to provide fair uncertainty results by the model output. The model output is influenced by the linearly added perturbations (derived from gradients) to its relevant input. Linear perturbations are the simplest and efficient way to reach the decision boundary in a small range of epsilon. Thus, the evenly spaced and symmetric interval enforces the inputs to move back and forth in an equal fashion with respect to the decision boundary, without any biasedness to any of the opposite class distributions.

### 4.2 Relation to Bayesian Approach

In fact, the proposed method can be viewed as a special case of an approximated Bayesian model. Under required conditions, applying prior to weights in a neural network can be matched with adding perturbations to the inputs. However, this is feasible on a very small interval of $\epsilon$ in a neighbourhood of zero. Let $y^*$ be the observed output corresponding to the input $x^*$ with $M$ dimensions. Then, the mapping function of a neural network can be written as:

$$p(y^*|x^*, X, Y) = \int p(y^*|x^*, w) p(w|X, Y) dw, \qquad (3)$$

where $p(y^*|x^*, w)$ is the likelihood of the model and $p(w|X, Y)$ is the posterior distribution among its weights. We can write an approximation of the model's weight underlying the posterior distribution as: $\hat{w}_{ij} = w_{ij} + \delta w_{ij}$, where $\delta w_{ij}$ is a small perturbations on the model's parameter which can be equivalent of adding priors to the weights.

For simplicity, consider as an example a neural network with a single output layer. Then the predictive posterior on each neuron $c_j$ can be expressed in its canonical form as:

$$c_j = \sum_{k=1}^{M} x_k . \hat{w}_{kj} = \sum_{k=1}^{M} (x_k . w_{kj} + x_k . \delta w_{ij}), \qquad (4)$$

where $j = 1, \ldots, P$ with P being the number of output neurons. Using MC-AA method, the adversaries over an input are written as: $\hat{x}_i = x_i + \delta x_i$. Similarly, we plug the adversaries into the model's equation as follows:

$$c_j = \sum_{k=1}^{M} \hat{x}_k . w_{kj} = \sum_{k=1}^{M} (x_k . w_{kj} + \delta x_k . w_{kj}) \qquad (5)$$

Assuming Eq. 4 is equivalent to 5, we obtain the necessary constraints:

$$\frac{\delta x_k}{x_k} = \frac{\delta w_{kj}}{w_{kj}} \qquad (6)$$

Generally, the condition in Eq. 6 safely holds true on very small perturbations. In other words, we can guarantee the satisfaction of this constraint by choosing $\frac{\delta x_k}{x_k}$ in a neighbourhood of zero. Consequently, this is true when $\epsilon$ is very small.

## 5 Model Uncertainty Using MC-AA

### 5.1 Obtaining and Evaluating Model Uncertainty

Firstly, the different values of $\epsilon$ introduce multiple outputs per instance. Hence, the predictive mean can be written as:

$$p_{MC-AA}(y^*|x^*) \approx \frac{1}{T} \sum_{i=1}^{T} \hat{y}^*\left(x^*_{\{\varepsilon_i\}}, W_1, \ldots, W_L\right), \tag{7}$$

where $x^*_{\varepsilon_i}$ is the perturbed $x$ by FGSM method associated to $\varepsilon_i$ with $T$ being the length of the discrete interval $I$ as introduced in Sect. 4, with $\varepsilon_{\max}$ is a hyper-parameter to be tuned.

Referring to [4], mutual information (MI) is shown to be an effective measurement of uncertainty prediction. Also, it is sometimes referred to epistemic uncertainty [20]. MI reflects the amount of information of the multiple outputs on each input. Thus, the uncertain prediction requires more amount of information by the model and consequently produces higher MI. Following the same procedure as in [4], MI can be expressed as:

$$\hat{I}(y^*|x^*, w) = \hat{H}(y^*|x^*, w)$$
$$+ \sum_c \frac{1}{T} \sum_{i=1}^{T} p(y^* = c|x^*_{\varepsilon_i}, w) \log p(y^* = c|x^*_{\varepsilon_i}, w), \tag{8}$$

where $c$ is the class label, and

$$\hat{H}(y^*|x^*, w) = -\sum_c p_{MC-AA}(y^* = c|x^*, w) \log p_{MC-AA}(y^* = c|x^*, w) \tag{9}$$

The uncertainty of the model can be evaluated using the measurements of a binary classification problem. In this paper, the uncertainty evaluations provide a similar approach of using Area-under-Curve (AUC) score and Receiver-Operation-Curve (ROC). However, these metrics are accompanied by the ground truth of labels besides the predictive uncertainty derived from MI as evaluated in [23]. Meaning, the output measurements of uncertainty estimates are derived from correct/incorrect classification tied with certain/uncertain predictions. As a result, we can distinguish between four possible states regarding model uncertainty measurements as follows:

- TN: Correct and certain: The predictions match the labels and the mutual information is low.
- FP: Correct and uncertain: The predictions match the labels, but the mutual information is high.
- FN: Incorrect and certain: The predictions do not match the labels, but the mutual information is low.
- TP: Incorrect and uncertain: The predictions do not match the labels and the mutual information is high.

The uncertainty mapping (certain/uncertain) is derived from the predictive uncertainty measurement. After computing these measurements, we set an uncertainty threshold, $T_u$. This threshold moves between the minimum and the maximum predictive uncertainty estimates in the test set. The correct/incorrect and certain/uncertain maps correspond to a new binary classification task. From the abovementioned states, true positive (TP), false positive (FP), true negative (TN), and false negatives (FN) values reflect the performance of the uncertainty classification task. These terms yield the following expressions that reflect the goodness of uncertainty estimates:

- Accuracy of uncertainty estimation:

$$\text{Accuracy} = \frac{\text{TN} + \text{TP}}{\text{TN} + \text{TP} + \text{FN} + \text{FP}}$$

- Negative Predictive Value (NPV): we desire that if the model is certain about its prediction, the prediction is assumed to be correct. This can be reformulated as conditional probability:

$$p(\text{correct}|\text{certain}) = \frac{p(\text{correct, certain})}{p(\text{certain})} = \frac{\text{TN}}{\text{TN} + \text{FN}}$$

which is equivalent to NPV ratio in the binary test.

- True Positive Rate (TPR): If the model is incorrect about its predictions, we desire to be uncertain about the predicted value. This is expressed as a conditional probability:

$$p(\text{uncertain}|\text{incorrect}) = \frac{p(\text{uncertain, incorrect})}{p(\text{incorrect})} = \frac{\text{TP}}{\text{TP} + \text{FN}}$$

which is equivalent to "recall" used in the binary test.

The abovementioned metrics reflect a good uncertainty performance of the model where better performance corresponds to the higher values of these metrics. These metrics are studied according to the variation of the uncertainty threshold $T_u$ As the threshold variates between the min/max uncertainty values, ROC and AUC are hence computed, in which we can easily derive a summarisable performance of the metrics. Instead of changing the threshold between min/max uncertainties, we normalise this threshold with respect to the uncertainty, where $T_u \in [0, 1]$. Hence, the expression of the threshold can be written as: $T_u = \frac{u - u_{\min}}{u_{\max} - u_{\min}}$ where $u_{\max}$ and $u_{\min}$ correspond to the maximum and minimum values of the predictive uncertainty measurement over the data, respectively.

### 5.2 Behaviour of MC-AA

To illustrate the behaviour of MC-AA, we generate 2D synthetic data of 12,000 instances by sampling from random normal distribution on each dimension with a common standard deviation of 0.25 and averages of values 0 and 1 corresponding to classes 0 and 1, respectively, referring to Fig. 1. We fit this data into a neural network comprised of two hidden layers with 20 neurons each chosen arbitrarily. The outputs are then squashed with the softmax function to produce the output classes. We use the NLLLoss function and Adam optimiser, then we arbitrarily set the learning rate to 0.01 and the number of epochs to 100. We assume that all labels belong to class 0. Hence, we apply MC-AA method using different values of $\epsilon$ that belongs to a symmetric interval with $\varepsilon_{\max} = -\varepsilon_{\min} = 5 \times 10^{-3}$ chosen arbitrarily, wherein the interval values are evenly spaced by $\beta$ which is arbitrarily set to $\frac{\varepsilon_{\max}}{10}$. Hence, only a few output samples are required to produce uncertainty estimates using MI. Referring to Fig. 3, we highlight the captured uncertainty in green, using different thresholds $T_u$. This method behaves to some extent as MC-dropout that captures epistemic and aleatoric uncertainty of the data points near the decision boundary [14]. These data points lack the knowledge about the model's decision. Since the perturbations directly influence the input data, MC-AA is more robust to wrong labels lying between the overlapping regions (aleatoric uncertainty) in which the recent uncertainty methods have failed to detect them. This will be further clarified in the next sections.

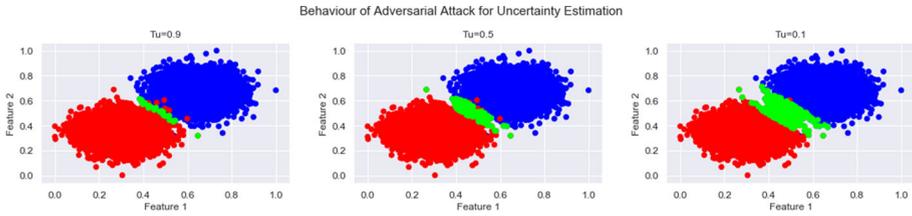

**Fig. 3** Representation of MC-AA behaviour on 2D synthetic data. Each subplot highlights the uncertain data points that fall above the provided threshold $T_u$

## 6 Experiments

To validate our method, we choose as an example two datasets derived from Bitcoin and Ethereum blockchain. We test the performance of MC-AA on these datasets. Then, we provide a fair comparison between our method (MC-AA) with the previous methods such as MC-dropout, DUQ and DUE using a same experimental set-up on each of the datasets. In these experiments, we use Pytorch [24] and GPytorch [25] packages in Python programming language. We further divide this section into two parts, wherein we test the proposed method with Bitcoin and Ethereum datasets in the first and second parts, respectively.

### 6.1 Experimenting with Bitcoin Data

It is known as Elliptic data which has been studied in [26–28] to detect illicit transactions. This data comprises 49 distinct transaction graph networks collected at 49 different timestamps, respectively. Each graph is directed acyclic graph (DAG) incorporating partially labelled nodes as transactions between Bitcoin addresses and the edges as the payments flow. The nodes acquire 166 features including timestamp, local features related to the transactions (e.g. transaction fees) and aggregated features derived from the one-hop aggregations with respect to the node of interest. Out of labelled nodes, 42,019 nodes are associated with licit (e.g. miners) and 4545 belong to illicit transactions (e.g. scams). Regarding data split, the primary 34 graphs correspond to the train set, and the 15 remaining graphs correspond to the test set, whereas the last 5 graphs of the train set are used as validation set. Afterwards, we exclude the timestamp feature and standardise the dataset in our experiments. The nodes of each timestamped graph are considered as a batch in this experiment. Using this data, we apply MC-AA, MC-dropout, DUQ and DUE each using a similar neural network as a base network consisting of two hidden layers of widths $n_1 = 100$ and $n_2 = 81$, chosen empirically, and squashed by ReLU function. However, each of the uncertainty methods is associated with its corresponding experimental settings, which is provided in what follows.

**MC-AA:** With MC-AA method, the base network is followed by the softmax output layer to produce the binary licit/illicit predictions. We empirically assign 0.3/0.7 to the weights of the NLLLoss function to mitigate the class imbalance. We use Adam optimiser with a learning rate that is empirically set to 0.01. After training the model, we perform MC-AA on the test set to capture the model uncertainty using multiple forward passes. Beforehand, all test points are assumed to be in the licit class. Using the best AUC-score of model uncertainty on the validation set, the hyper-parameter $\varepsilon_{\max}$ is empirically tuned to be 0.1. We limit the number of samples per data point to 20 by setting $\beta$ arbitrarily to $\frac{\varepsilon_{\max}}{10}$. Then, the predictive uncertainty is obtained using MI over the provided samples.

**MC-dropout:** Like the preceded model in MC-AA, we use a similar neural network but with a dropout function after every hidden layer. We empirically choose dropout equals 0.3. Then, we arbitrarily perform 50 stochastic forward passes on the test set to produce uncertainty estimates using MI.

**DUQ:** DUQ model comprises of feature extractor and an additional learnable layer known as RBF kernel with two-sided gradient penalty to produce reliable uncertainties with single forward pass [5]. Here, the feature extractor is the given base network to produce a fair comparison. The output layer (RBF kernel) is formed of a learnable weight matrix per class in order to compute the distance from the data points to the centroids of the different classes. So, the weight matrix embeds the output of the feature extractor in a new space of embedding size (centroid size) set by default to 10. The centroid of each class is updated by an exponential moving average of the features corresponding to data points in accordance with that class. The gradient penalty is the $l_2$ norm of the gradient's output with respect to the input bounded by Lipschitz constant of 1 as introduced in [5]. This penalty incorporates a regularisation factor $\lambda$. The binary-cross-entropy loss function is used to minimise the distance to the correct centroid and maximise the others. Using Bitcoin data, we empirically set the hyper-parameters of DUQ as : $\sigma$ of RBF kernel equals 0.3, $\lambda = 0.1$, $\gamma$ as a factor of moving average over centroids is set to 0.9 and learning rate equals to 0.001.

**DUE:** DUE model comprises a feature extractor that is followed by Gaussian Process (GP) [6]. We use the same feature extractor as preceded. The model arbitrarily utilises 15 inducing points that are initialised equally to the centroids by clustering the feature extractor output using Kmeans algorithm with 15 clusters. Then, the output of the feature extractor is fed to a sparse GP kernel with the variational inducing points that are used to approximate the full dataset by maximising a lower bound on the marginal likelihood known as ELBO (evidence lower bound). This model produces various predictions using the softmax likelihood function by a single forward pass, wherein the predictive uncertainty is then computed via MI. Using Bitcoin data, we empirically obtain the hyper-parameters settings as follows: learning rate = 0.01 and epochs = 50.

### 6.2 Experimenting with Ethereum Data

This data consists of labelled non-fraud/fraud accounts with a valid transaction history over Ethereum blockchain which is studied in [29]. The fraudulent accounts are acquired by Ethereum community to detect illicit behaviour e.g. Ponzi schemes and scams. Ethereum account data comprises 9841 labelled account with 77.86% as non-fraud and 22.14% as fraud. Initially, this dataset has an overall of 49 numerical and categorical features e.g. "Total ether sent" and "total number of sent/received transactions". We exclude categorical features as there are missing values. Also, we perform data pre-processing to exclude features with zero variance or whose correlations are higher than 0.9, chosen arbitrarily. Then, we remove three more features that are formed of less than 10 unique values, which are experimentally found to be non-informative on the classification. Thus, the feature space is reduced to 28 dimensions and the data is randomly split to train/validation/test by the ratios 0.7/0.1/0.2, wherein the features are standardised. With Ethereum dataset, we apply MC-AA, MC-dropout, DUQ and DUE each using neural network as a base network consisting of two hidden layers of widths $n_1 = 50$ and $n_2 = 25$ chosen empirically, and squashed by ReLU function. In the following parts, we provide the necessary experimental set-up on each of the uncertainty methods.

**MC-AA:** We apply the softmax function to output the non-fraud/fraud predictions on the output layer of the mentioned base network. We use the weighted NLLLoss with weights 0.4/0.6, chosen empirically. The number of epochs and the learning rate are empirically set to 50 and 0.01, respectively. The batch size is arbitrarily set to 512. Using MC-AA on the trained model, we assume that all testing points belong to class non-fraud. We empirically choose the hyper-parameter $\epsilon_{max}$ equals to $8.1 \times 10^{-4}$ that provides the highest AUC-score on the validation set. We set the number of samples per data point to 20 by setting $\beta$ arbitrarily to $\frac{\varepsilon_{max}}{10}$.

**MC-dropout:** Similarly to the preceded model in MC-AA, we apply an additional dropout function after every hidden layer of empirically chosen dropout ratio equals 0.3. Then, we perform 50 stochastic forward passes on the test set, in which the uncertainty estimates are computed through MI.

**DUQ:** We use the base network as feature extractor for DUQ. For this dataset, we empirically opt the following values for hyper-parameters: epochs $= 50$, learning rate $= 0.001$, $\sigma = 0.3$, $\lambda = 0.01$, $\gamma = 0.9$ and batch size $= 512$.

**DUE:** With Ethereum data, we use the base network as feature extractor and we empirically choose the following settings: epochs $= 50$, learning rate $= 0.01$ and batch size $= 512$.

## 7 Discussion

Regarding Bitcoin data, we plot and compare the different uncertainty measurements on the above-mentioned methods as shown in Figs. 4 and 5. Remarkably, the model uncertainty based on MC-AA has shown superior performance in comparison to other uncertainty methods. Referring to Fig. 4, NPV and TPR measurements have revealed a remarkable improvement using MC-AA, whereas the accuracy curve has shown acceptable but lower performance. This means that we are able to capture more true positives (incorrect and uncertain) and less false negatives (incorrect and certain) on the cost of having more false positives (correct and uncertain). This scenario is more favoured in model uncertainty because being certain on incorrect predictions affects the integrity of the model. While, the uncertain and correct predictions can be forwarded to an annotator for further analysis. In addition, the goodness of classification of model uncertainty has recorded the highest AUC-score equals to 0.8 with MC-AA, whereas others have recorded AUC-score of less than or equal to 0.75.

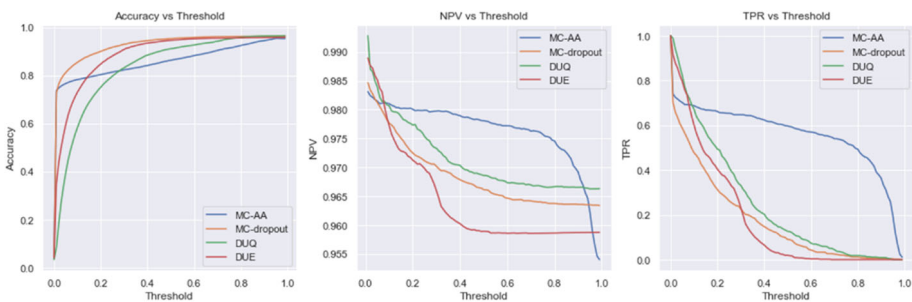

**Fig. 4** Comparison of model uncertainty using different uncertainty methods on Bitcoin data. The subplots (from left to right) correspond to accuracy, NPV and TPR as a function of threshold $T_u$

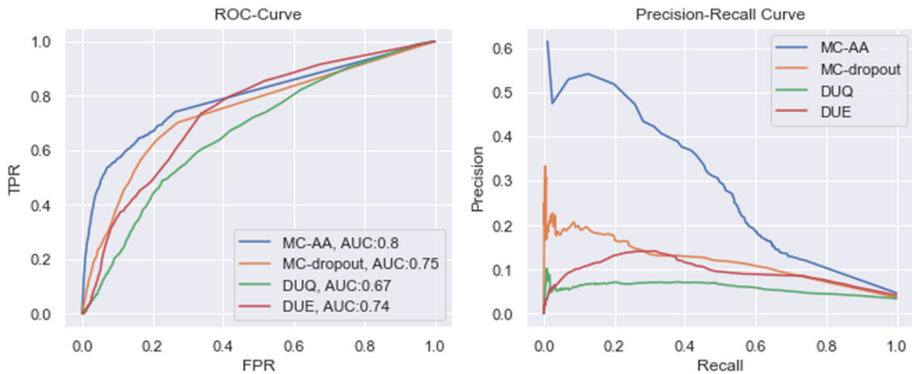

**Fig. 5** ROC-curve and Precision-Recall as a function of threshold $T_u$ of model uncertainty using Bitcoin data. The model uncertainty is performed using MC-AA (Ours), MC-dropout, DUQ and DUE

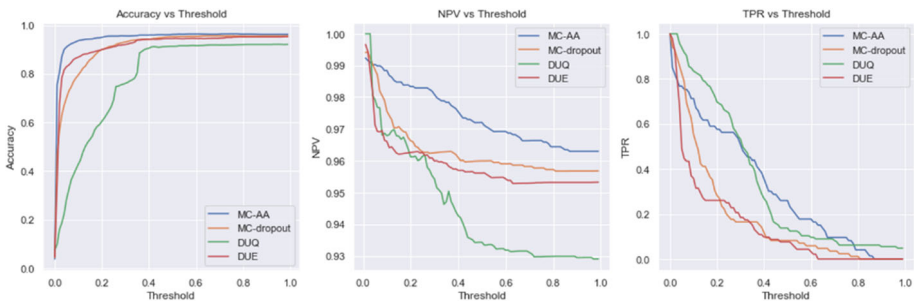

**Fig. 6** Comparison of model uncertainty using different uncertainty methods on Ethereum data. The subplots (from left to right) correspond to accuracy, NPV and TPR as a function of threshold $T_u$

Also, Precision-Recall curve has admitted a significant outperformance using our proposed method as shown in Fig. 5.

In Ethereum data, we follow the same procedure to evaluate and validate our method in comparison to previous uncertainty methods. Referring to Figs. 6 and 7, the model uncertainty based on MC-AA has outperformed the other mentioned methods with an AUC-score equals to 0.88.

The performance of MC-AA method is tied with the tuned hyper-parameter $\varepsilon_{max}$ It is feasible to force the "incorrect and certain predictions" to be lower by increasing the value of this hyper-parameter, but at the cost of other measurements. MC-AA can be also viewed as a generalised form of the standard neural network model and can be approximated with MC-dropout for a certain dropout value and $\varepsilon_{max}$. However, MC-AA has shown to be more favoured than MC-dropout since we have higher guidance and influence on the uncertainty performance in accordance to the classification problem. For instance, the changes in $\varepsilon_{max}$ of MC-AA are tied with perturbed inputs in which we can force the model not to tolerate data points lying in the wrong class. While in MC-dropout, the changes in dropout factor are tied with the perturbation of the decision boundary, wherein the model cannot be uncertain about data points in the wrong class. In general, MC-AA is a viable approach in a binary classification tasks and not limited to the used datasets. On the other hand, dealing with multi-class classification is not applicable with this method unless the classifier is converted into "one versus all" classification. According to the results, MC-AA has shown to be more efficient than MC-dropout as our proposed method has produced uncertainty estimates with 10 forward passes whereas the latter method has performed 50 forward passes to provide stable uncertainty estimates using mutual information. Meaning, we are able to achieve good uncertainty estimates with 10 back and forth movements using MC-AA.

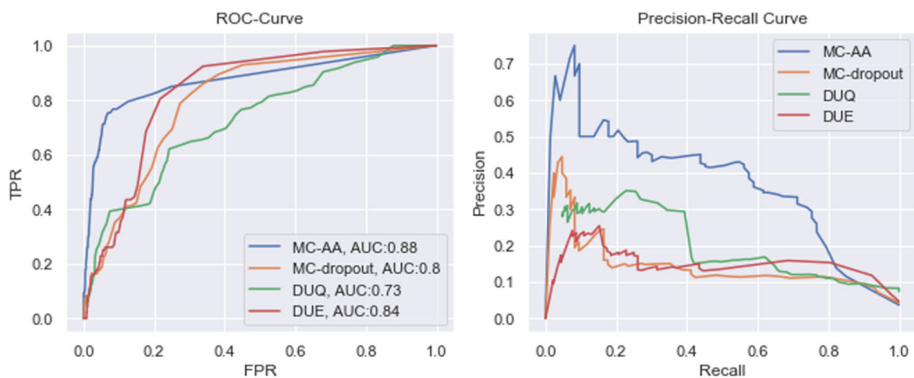

**Fig. 7** ROC-curve and Precision-Recall as a function of threshold $T_u$ of model uncertainty using Ethereum data. The model uncertainty is performed using MC-AA (Ours), MC-dropout, DUQ and DUE

## 8 Conclusion

We have proposed a novel method (MC-AA) based on adversarial attacks to capture model uncertainty in binary classification tasks. We have shown that this method provides reliable uncertainty estimations with reduced number of misclassified and certain predictions. We have compared the performance of MC-AA with previous methods. Our model has outperformed the recent studies using datasets derived from Bitcoin and Ethereum blockchain.

**Publisher's Note** Springer Nature remains neutral with regard to jurisdictional claims in published maps and institutional affiliations.